\documentclass[a4paper,twoside]{article}

\usepackage{stfloats}
\usepackage{epsfig}
\usepackage{subcaption}
\usepackage{calc}
\usepackage{amssymb}
\usepackage{amstext}
\usepackage{amsmath}
\usepackage{amsthm}
\usepackage{multicol}
\usepackage{pslatex}
\usepackage[bottom]{footmisc}
\usepackage{SCITEPRESS}     

\usepackage[
   sorting=nyt, 
   citestyle=authoryear-comp
   maxcitenames=2,
   maxbibnames=20,
   natbib=true,
   backend=bibtex, 
   style=authoryear-comp,
   sorting=none,
   isbn=false,
   url=false,
   doi=false,
   eprint=false,
   giveninits=true, 
]{biblatex}
\bibliography{Example} 
\begin{document}

\title{MixedTeacher :  Knowledge Distillation for fast inference textural anomaly detection}

\author{Simon Thomine$^{1,2}$, Hichem Snoussi$^{1}$, Mahmoud Soua$^{2}$\\
	\normalsize $^{1}$University of technology Troyes, Troyes, France \\
	\normalsize $^{2}$ AQUILAE, Troyes, France \\
	\normalsize e-mail: simon.thomine@utt.fr, hichem.snoussi@utt.fr, m.soua@aquilae.tech
}

\keywords{Anomaly Detection - Texture - Knowledge Distillation - Layer Selection - Unsupervised}

\abstract{For a very long time, unsupervised learning for anomaly detection has been at the heart of image processing research and a stepping stone for high performance industrial automation process. With the emergence of CNN, several methods have been proposed such as Autoencoders, GAN, deep feature extraction, etc. In this paper, we propose a new method based on the promising concept of knowledge distillation which consists of training a  network (the student) on normal samples while considering the output of a larger pretrained network (the teacher). The main contributions of this paper are twofold: First, a reduced student architecture with optimal layer selection is proposed, then a new Student-Teacher architecture with network bias reduction combining two teachers is proposed in order to jointly enhance the performance of anomaly detection and its localization accuracy. The proposed texture anomaly detector has an outstanding capability to detect defects in any texture and a fast inference time compared to the SOTA methods. }

\onecolumn \maketitle \normalsize \setcounter{footnote}{0} \vfill

\section{\uppercase{Introduction}}
\label{sec:introduction}

Anomaly detection in industry is a vast topic since there is a lot of possible applications. For instance, defect detection aims at identifying specific anomaly classes and locations in industrial manufacturing processes \citep{kahler_anomaly_2022}. This detection is crucial for ensuring the high quality of final products \citep{minhas_anonet_2019}. A common property of defects is that their visual texture is inherently different from the defect-free surface. The specificity of textures is the pattern structure which, if known, allows the detection and the extraction of anomalies. However, the texture anomaly generally appears in a small region in few samples, which makes it difficult to build consistent normal and abnormal datasets to be used in supervised learning methods. Hence, unsupervised anomaly detection networks are very suitable for industrial scenarios as they represent the strong basis for building a detection model without any annotated samples \citep{huang_unsupervised_nodate}. Several unsupervised anomaly detection methods have been introduced for texture anomaly detection. These methods could achieve high performance up to 99.6 AUROC. However, they suffer from complex networks and high latency.\\ 

\noindent In another context, knowledge distillation has been introduced with the purpose of reducing the network size while increasing performance. Knowledge distillation aims to train a smaller network (student) to imitate pretrained one or several larger ones (teachers) on normal samples. As the teacher is pretrained, it has the ability to generalize even if the sample contains an anomaly, whereas the student won't be able. Hence, by comparing the extracted features between the teacher and the student networks, an abnormal sample could be detected. According to some studies \citep{iglesias_analysis_2015}, using too many features can significantly reduce the accuracy of anomaly detection. Recently, a Student-Teacher Feature Pyramid Matching (STPM) method has been proposed in \citep{wang_student-teacher_2021}, where the first three network layers are used in order to focus on edges, colors and shapes instead of context information. Even if using layer selection technique is an interesting approach, there is still a lack of explanation concerning the layer choice and the relevance of the relative information. Looking at the same layers for an object and for a texture reduces the relevance of the extracted information. For example, looking at context information in a texture is pointless and for an object, pure edge/color/texture information may not be the most interesting information.  
		
\noindent Another recurrent problem is the classifier bias. The best current methods use a pretrained classifier network on imageNet which is biased by the classes of imageNet and can have an impact on the localization and the detection of defects. \\

\noindent The main contributions of the paper are as follows: 
\begin{itemize}
\item A new reduced student architecture for texture-specific object category.

\item In order to reduce the classification bias, we propose a new architecture combining two teachers pretrained on imageNet but with different architectures (respectively ResNet-18 \citep{he_deep_2015} and EfficientNet-b0 \citep{tan_efficientnet_2020}) and a single student network. This new mixed Teacher network structure outperforms competitive state-of-the-art methods both in inference time and SOTA scores, on anomaly datasets such as MVTEC AD textures and BTAD textures \citep{mishra_vt-adl_2021}. The proposed MixedTeacher model uses a score and anomaly localisation function based on each complementary teacher features with a careful feature selection.

\end{itemize}

The paper is organized as follows. In section 2, we review the related work, especially on MVTEC dataset and present the different approaches proposed in literature. In section 3, we compare the results of training with different architectures and different layer selection schemes and introduce our proposed texture-specific reduced student architecture. Section 4 is dedicated to describing a novel mixed Student-Teacher network. In section 5, we compare our results to the SOTA methods for both the reduced student architecture and the MixedTeacher in terms of AUROC, pixel-AUROC and inference time.

\section{\uppercase{RELATED WORK}}

 Anomaly detection is a problem that pops up in many areas and is often very difficult to deal with. Indeed, detecting the ``abnormal" is a rather vague concept and is difficult to define according to the use cases, which makes research on this subject very specific.
 
For several years, the rise of deep learning has never ceased to impress with high quality results and interesting methods. Most of these methods are based on an unsupervised representation approach to discriminate outliers. Some specific work has been done for fabrics defect detection such as the multi-scale Convolutionnal denoising autoencoder \citep{mei_automatic_2018}. For unsupervised anomaly detection in general, we can also cite the GEE, a gradient based VAE \citep{nguyen_gee_2019} or the Gaussian mixture model VAE \citep{nguyen_gee_2019}. Another common way to detect anomaly is to use generative adversarial networks \citep{goodfellow_generative_2014}. Ano-GAN \citep{schlegl_f-anogan_2019} was one of the first utilization of GAN for anomaly detection but since then a lot of approaches emerged such as G2D \citep{pourreza_g2d_2021} and OCR-GAN \citep{liang_omni-frequency_2022}. Other interesting approaches rely on pretrained models especially on imageNet, using the feature extraction of pretrained network to extract useful information about a given sample. The idea is to extract features with a pretrained model and then train a normalizing flow model on good samples, so that the model is ready to find out if a given sample is an anomaly by looking at the reconstruction error. An advantage of normalizing flow is the reversible aspect which is useful to locate the anomaly pixel-wise. Many techniques based on this concept have been proposed such as differNet \citep{rudolph_same_2021} and CS-FLOW \citep{rudolph_fully_2021} which consider multi-scale normalizing flow and FastFlow \citep{yu_fastflow_2021} based on a 2D normalizing flow. 

Recently, the concept of knowledge distillation has also been used for unsupervised anomaly detection. The student-teacher method consists of training a student teacher based on the output of a larger teacher model which is pretrained on ImageNet. The student network will learn to imitate the teacher on good samples only. Then, when an abnormal sample is tested, the teacher will be able to generalize and the student won't be, the difference between the output of the teacher and the output of the student will allow the detection of the anomaly. On the MVTEC dataset, four methods have been implemented, STPM \citep{wang_student-teacher_2021} which trained the student on the 3 first layers of ResNet-18, RSTPM \citep{yamada_reconstruction_nodate} which is basically the same method but with an attention layer, reverse distillation \citep{deng_anomaly_2022} and CFA \citep{lee_cfa_2022}.

\section{LAYER SELECTION AND REDUCED STUDENT}
In this section, after a comparative study of  layer selection methods for optimal texture anomaly detection, we present a new student architecture that both increases performance and reduces the inference time.
\subsection{Layer selection} 
 In deep neural networks, a common observation is that deep layer features contain context information and shallow layer features contain color, texture and contour information. In a case of detection of defects on the fabric or on a generic texture, the context information is less important than the texture information, therefore, we will turn to shallow layer features. As reported in table \ref{table:Table1}, different combinations of shallow layers have been tried in order to select the optimal architecture with respect to detection performance evaluated by the AUC.
\begin{table}[h]
	\centering
	\caption{\textbf{Layers selection results}}
	\footnotesize
	\setlength{\tabcolsep}{9pt}
	\renewcommand{\arraystretch}{1.3}
	\scalebox{0.8}{
	\begin{tabular}{|c|cc|}
			
		\hline
		{\centering \textbf{Measures}} & {\textbf{Layer 1 and 2 AUC}} &
		{\textbf{Layer 2 and 3 AUC}} \\
	
		\hline
		{Mean objects} &
		{\centering 0.876}  &
		{\centering 0.910}  \\
		{Mean textures} &
		{\centering 0.990}  &
		{\centering 0.971}  \\
	    \hline
 		
	\end{tabular} \label{table:Table1}
	}
\end{table}

\subsection{Reduced student} 
\begin{figure*}[b]
\centerline{\includegraphics[scale=0.3]{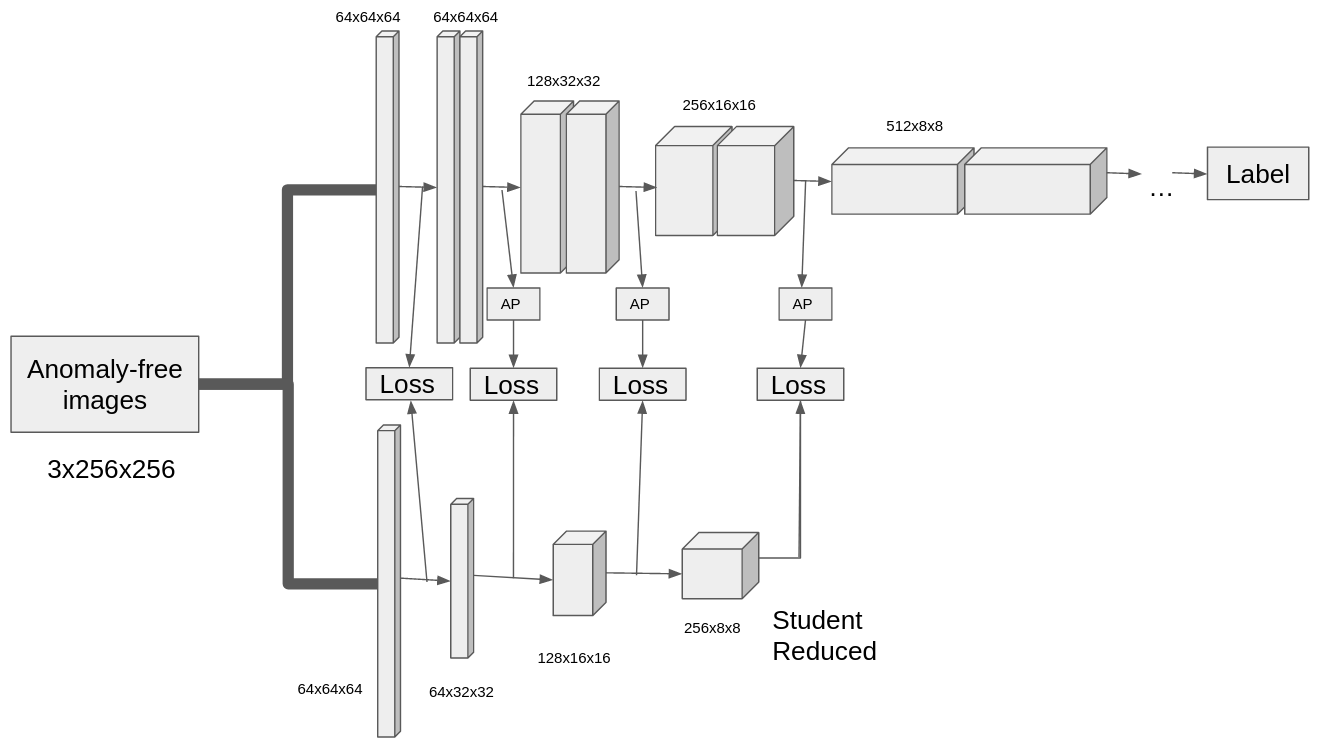}}
\renewcommand{\arraystretch}{1}
\caption{ 
Reduced student architecture with AP for adaptive average pooling. }
\label{Fig.1}
\end{figure*}

 ResNet-18 architecture has been retained for the teacher network. As  texture specific anomaly detection is the main objective of this work, we propose to add the ResNet-18 first layer after the first convolution to extract even more textural information. The second objective was to alleviate the student architecture to decrease inference time and possibly performance. As ResNet-18 presents several residual blocks with two identical convolutional layers, we first decided to take only one layer for each block in our student architecture. The classifier bias is another known problem while dealing with pretrained classifier and we tackled this problem by reducing features size with an adaptive average pooling layer in each Resnet residual block's output as presented in figure \ref{Fig.1}.\\

\noindent Given a training dataset of images without anomaly ${D=[I_1,I_2,...,I_n]}$, our goal is to extract the information of the $L$ bottom layers. For an image ${I_k} \in R^{w*h*c}$ where $w$ is the width, $h$ the height and $c$ the number of channels, the teacher outputs features $F_t^l(I_k) \in R^{w_l*h_l*c_l}$ and $F_s^l(I_k) \in R^{w_l/2*h_l/2*c_l/2}$ with $l>1 $ and $F_s^l(I_k) \in R^{w_l*h_l*c_l}$ if $l=1$. For the loss function, we took the $l2$ distance of normalized feature vectors like in the STPM original paper \citep{wang_student-teacher_2021} while using an adaptive average pooling on teacher features where $l>1$ and just sum all feature maps of all layers to obtain our loss with the same ratio for all layers (Eq.\ref{eq.1}).

\begin{equation}
F_t^{l>1}(I_k)=AAP(F_{Resnet18}^{l>1}(I_k))
\label{eq.1}
\end{equation}

\noindent where AAP refers to the Adaptive Average Pooling. Pixel loss is defined in the following Eq.\ref{eq.2}: 

\begin{equation}
loss^{l}(I_k)_{ij}=\frac{1}{2}\lVert norm(F_t^l(I_k)_{ij})-norm(F_s^l(I_k)_{ij}) \rVert 
\label{eq.2}
\end{equation}

\noindent and for the layer l, the loss is defined as: 
\begin{equation}
loss^{l}(I_k)=\frac{1}{w_lh_l}  \sum_{i=1}^{w_l} \sum_{j=1}^{h_l} loss_{resNet}^l(I_k)_{ij}  
\label{eq.3}
\end{equation}

\noindent  and finally for the total loss is written as: 

\begin{equation}
loss(I_k)= \sum^{l} loss^{l}(I_k) 
\label{eq.4}
\end{equation}

Performance and inference speed are later reported in section 5 with comparison with SOTA networks on anomaly detection.

\section{MIXED TEACHER}
In this section, we introduce our new student teacher network structure that combines two teachers with the purpose of reducing the classifier bias, taking benefits from the two networks and exploiting the different layers in an optimal way.

\subsection{Observation and main ideas}
\begin{figure}[h]
\centerline{\includegraphics[scale=0.20]{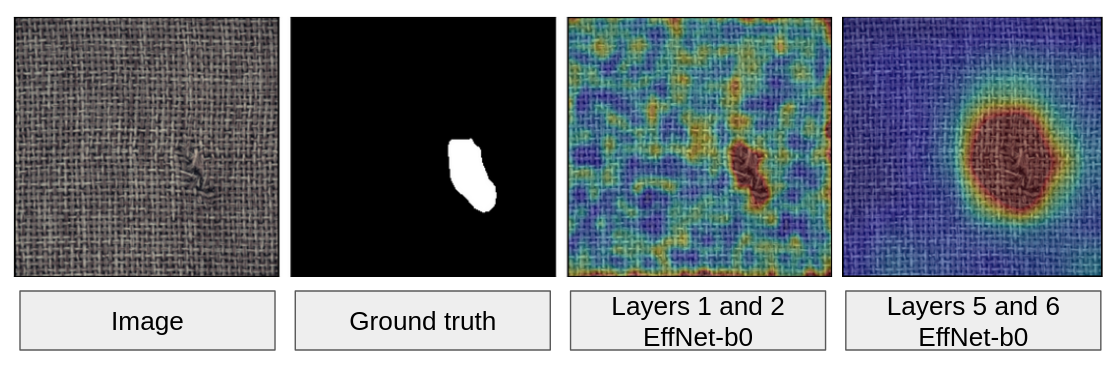}}
\renewcommand{\arraystretch}{1}
\captionsetup{justification=centering}
\caption{ 
Difference between top layers and bottom layers for EfficientNet-b0 architecture.}
\label{Fig.2}
\end{figure}

While testing our new student reduced architecture on the MVTEC AD textures, we obtained good results, but some noise still degrade results in terms of default localisation on specific images or texture-specific normal variation. Different teacher network architectures have been tested to conclude that ResNet-18 remains the best in terms of average precision and speed. However,  interesting behaviors have been observed on the noise localisation for each architecture. In fact, every classifier had the capacity to locate the anomaly, but with output noise and anomaly detection mistakes. \\
The combination of two pretrained classifier networks has therefore been proposed with the purpose of interpolating their defect localisation to cancel noise and false detection/segmentation. \\
EfficientNet-b0 has been proposed as the second teacher when considering its performance in terms of precision and speed. For this network, it has been observed that for the bottom layers, one has good localisation but with a huge noise and with top layers, a coarse defect localisation but with minimal noise has been obtained, as illustrated in figure \ref{Fig.2}.

\subsection{Method description}
\begin{figure*}[h]
\centerline{\includegraphics[scale=0.4]{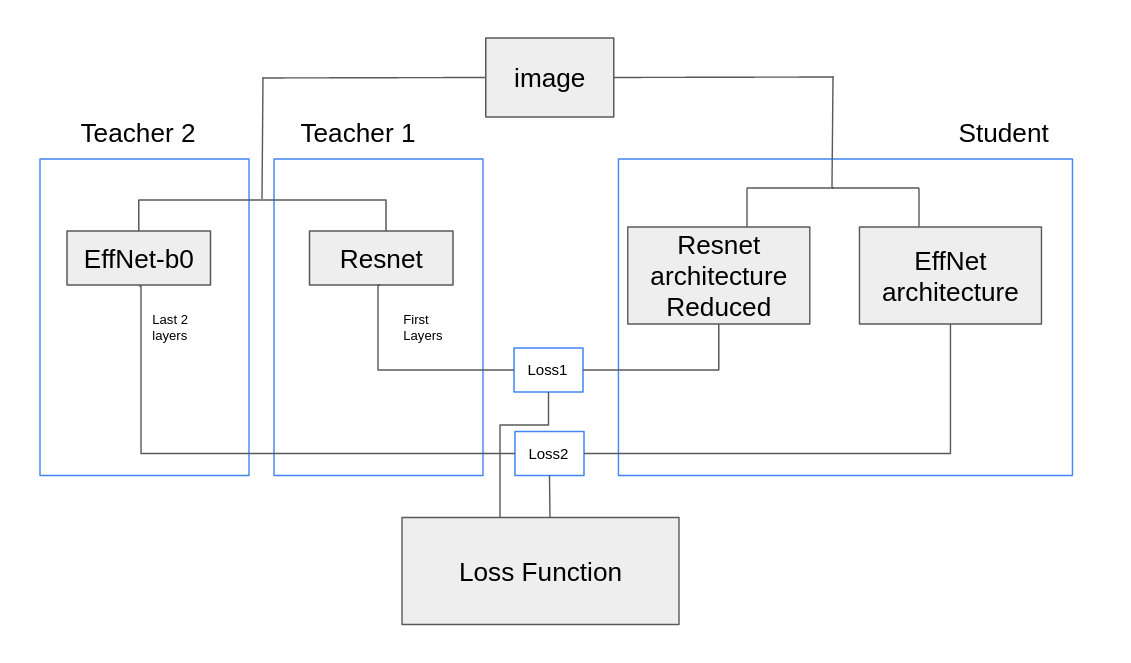}}
\renewcommand{\arraystretch}{1}
\caption{ 
MixedTeacher architecture. }
\label{Fig.3}
\end{figure*}

The learning architecture is composed of two teachers: the ResNet-18 as main teacher and EfficientNet-b0 as a localisation confirmation teacher. For the ResNet-18 part, the reduced student proposed in section 3 is used in order to ensure a good inference speed and precision on texture samples. For EfficientNet-b0 student, we used one convolution for each efficientnet block without pooling because we used deepest layers. In the student architecture, there is no communication between the networks except for the loss function as shown in figure \ref{Fig.3}.

For the training loss function, we used basically the same loss function as the one for the reduced teacher and we add an $\alpha$ factor to smooth the layer activation difference from the two teacher networks. As feature difference in efficientNet was about 10 times bigger than in ResNet-18,  $\alpha$ has been set to 0.1.

\begin{equation}
loss_{effNet}^{l=5,6}(I_k)_{ij}=\frac{1}{2}\lVert norm(F_t^l(I_k)_{ij})-norm(F_s^l(I_k)_{ij}) \rVert
\label{eq.5}
\end{equation}

and 

\begin{equation}
loss_{effNet}^{l=5,6}(I_k)=\frac{1}{w_lh_l}  \sum_{i=1}^{w_l} \sum_{j=1}^{h_l} loss_{effNet}^l(I_k)_{ij}   
\label{eq.6}
\end{equation}

\noindent and for Resnet-18 part : 

\begin{equation}
loss_{resNet}^{l=1,2,3}(I_k)=\frac{1}{w_lh_l}  \sum_{i=1}^{w_l} \sum_{j=1}^{h_l} loss_{resNet}^l(I_k)_{ij}  
\label{eq.7} 
\end{equation}

\noindent with $loss_{resNet}^{l}(I_k)_{ij}$ defined as in section 3. For the total loss with the $\alpha$ factor : 

\begin{equation}
loss_{tot}(I_k)= \sum_{l=1}^{3} loss_{resNet}^{l}(I_k) + \alpha \sum_{l=5}^{6} loss_{effNet}^{l}(I_k)
\label{eq.8}
\end{equation}

As in every knowledge distillation method, the loss only impacts the student.

\subsection{Anomaly score and localisation}

In the test phase (inference), we want  an anomaly map $M$ of the original image size where every pixel at
position $(i,j)$  has an anomaly score $M_{ij}$. With a test
image $I$ and $F_{tResnet}^l$ , $F_{tEffNet}^l$ the two teachers features
of $l$th layer and $F_{sResnet}^l$, $F_{sEffNet}^l$ their corresponding
$l$th layer student features, 
we perform an upsample on the difference between the corresponding layers. The
coarse localisation output of the efficientNet layers is
obtained by summing each layer’s anomaly map.

The anomaly map is obtained in the same way for the resnet part.
Respectively : 
\begin{equation}
A_{mapEffnet}= \sum_{l=5}^{6} Upsample(F_{tEffNet}^l-F_{sEffNet}^l)
\label{eq.8}
\end{equation}
and : 
\begin{equation}
A_{mapResnet}= \sum_{l=1}^{3} Upsample(F_{tResnet}^l-F_{sResnet}^l)
\label{eq.8}
\end{equation}

We then multiply the resnet anomaly map by the normalization of the effnet anomaly map multiplied by its mathematical extent. 
With $A_{mapEffnet}$ , the anomaly
map of efficientNet layers and $A_{mapResnet}$ the anomaly
map of resnet layers, the final anomaly map is then defined
as :

\begin{equation}
\begin{split}
M=A_{mapResnet}*(max(A_{mapEffnet})- \\ min(A_{mapEffnet}))A_{mapEffnet}
\end{split}
\label{eq.9}
\end{equation}

\noindent The anomaly score is defined as :

\begin{equation}
score=\sum_{i=1}^{w} \sum_{j=1}^{h} M_{i,j}
\label{eq.10}
\end{equation}

\noindent with $w$ and $h$ are respectively the width and height of the anomaly map.

\section{EXPERIMENTS}

\subsection{Datasets}
\begin{figure}[h]
\centerline{\includegraphics[scale=0.32]{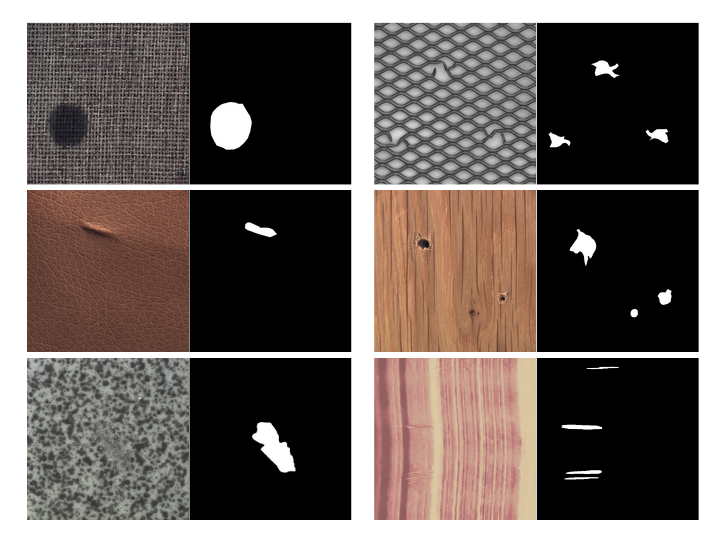}}
\renewcommand{\arraystretch}{1}
\captionsetup{justification=centering}
\caption{ 
Overview of textures from MVTEC AD and BTAD dataset, samples with anomaly and ground truth. These images are only used for testing and unseen during the training.}
\label{Fig.4}
\end{figure}

We experiment our methods on the textures from the  \textbf{MVTEC AD} \citep{bergmann_mvtec_2019} dataset which consists of 15 categories : 5 textures and 10 objects with a total of more than 5000 high resolution images. This dataset is used for unsupervised anomaly detection therefore it contains only anomaly free images for the training. For the test part, it shows a good variety of defects with ground truth masks for anomaly localisation. 
We also used the texture of the  \textbf{BTAD} \citep{mishra_vt-adl_2021} dataset which is an unsupervised anomaly dataset with three different categories including one texture figure \ref{Fig.4}.

The performance is evaluated with AUROC metric image-level and pixel-level to compare our results with other methods.

\subsection{Implementation and training metrics}
Training and inference were done on an rtx 2080ti. \\
To test the student reduced, we used the features of the first three blocks and the layer before the first block of ResNet-18. The Resnet network was pretrained on imageNet. We used stochastic gradient descent with a learning rate of 0.4 for 100 epochs with a batch size of 16. To test the MixedTeacher, we used the output features of the first two blocks and the layer before the first block of ResNet-18 and the output features of block 5 and 6 of EfficientNet-b0. We used stochastic gradient descent with a learning rate of 0.4 for 200 epochs with a batch size of 16. Both networks are pretrained on imageNet.
We resized all the images to a size of 256x256 keeping 80\% for training and 20\% for validation. We kept the checkpoint with the lowest validation loss.

\subsection{Reduced Student}
\subsubsection{Performance results}
In this paragraph, we will compare reduced student AUROC results to SOTA methods. In \ref{table:Table2}, we present AUROC performance results of CFA \citep{lee_cfa_2022}, PatchCore \citep{roth_towards_2021}, FastFlow \citep{yu_fastflow_2021}, STPM \citep{wang_student-teacher_2021}, CutPaste \citep{li_cutpaste_2021} and our reduced student on MVTEC AD textures.

\begin{table}[h]
	\centering
    \captionsetup{justification=centering}
	\caption{\textbf{Image-AUROC comparison on MVTEC AD : Reduced Student}}
	\footnotesize
	\setlength{\tabcolsep}{2pt}
	\renewcommand{\arraystretch}{1.3}
	\scalebox{0.9}{
	\begin{tabular}{|c|ccccc|c|}
			
		\hline
		{\centering \textbf{Category}} & 
		{\textbf{CutPaste}} &
		{\textbf{CFA}} &
		{\textbf{PatchCore}} &
		{\textbf{STPM}} &
		{\textbf{FastFlow}} &
		{\textbf{Ours}} \\
	
		\hline
		{carpet} &
		{\centering 100}  &
		{\centering 97.3} &
		{\centering 98.7}&
		{\centering 95.4}&
		{\centering 99.4}&
		{\centering 100} \\
			
    	{tile} &
		{\centering 100}  &
		{\centering 99.4} &
		{\centering 98.7}&
		{\centering 94.9}&
		{\centering 100}&
		{\centering 98.7} \\
		
		{grid} &
		{\centering 96.2}  &
		{\centering 99.2} &
		{\centering 98.2}&
		{\centering 98.2}&
		{\centering 100}&
		{\centering 99.7} \\
		
		{wood} &
		{\centering 99.1}  &
		{\centering 99.7} &
		{\centering 99.2}&
		{\centering 96.1}&
		{\centering 99.2}&
		{\centering 99.6} \\
		
		{leather} &
		{\centering 95.4}  &
		{\centering 100} &
		{\centering 100}&
		{\centering 98.9}&
		{\centering 99.9}&
		{\centering 99.7} \\
		
		\hline
		{Mean} &
		{\centering 98.1}  &
		{\centering 99.1} &
		{\centering 99.0}&
		{\centering 96.7}&
		{\centering 99.7}&
		{\centering 99.5} \\
		
	    \hline
 		
	\end{tabular} \label{table:Table2}
	}
\end{table}

For FastFlow, we choosed to take the results from Anomalib as we were not able to reproduce their paper results (99.9 AUROC in paper). As seen in table \ref{table:Table2}, our reduced student is better than CFA for texture anomaly detection, which is the best actual knowledge distillation unsupervised anomaly detection method and is close to the SOTA results. We manage to gain 2.8 points against classic STPM with a network reduction and a wise layer selection aiming for texture specific anomaly detection.

\subsubsection{Inference time results}
In table \ref{table:Table3}, we compare the reduced student inference time to other SOTA methods. The main purpose of reduced student was to propose a high processing speed to manage real time for several high resolution images. To get inference time results, we employ Anomalib. All the additional results come from this library to make sure the tests were carried out under the same conditions.

\begin{table}[h]
	\centering
	\caption{\textbf{Inference time results}}
	\footnotesize
	\setlength{\tabcolsep}{2pt}
	\scalebox{1}{
	\begin{tabular}{|c|ccc|c|}
			
		\hline
		{\centering \textbf{Category}} & 
		{\textbf{PatchCore}} &
		{\textbf{FastFlow}} &
		{\textbf{STPM}} &
		{\textbf{Ours}} \\
	
		\hline
		
		{FPS} &
		{\centering 5.8}  &
		{\centering 21.8} &
		{\centering 83.2}&
		{\centering \textbf{108.1}}\\
		{Latency (ms)} &
		{\centering 172}  &
		{\centering 45.9} &
		{\centering 12}&
		{\centering \textbf{9.2}}\\
		\hline
 		
	\end{tabular} \label{table:Table3}
	}
\end{table}
The presented results are based on Anomalib inference time. In a self made code, we were able to obtain a 10x better inference time for STPM and reduced student. The most important thing to consider is that the STPM is by far the fastest anomaly detector and reduced student reduced its inference time by 30\%.

\subsection{MixedTeacher}
\subsubsection{Performance results}
Unlike the reduced student, the MixedTeacher main purpose is performance and not inference time. In table \ref{table:Table4} we compared AUROC of several SOTA methods in texture anomaly detection.

\begin{table}[h]
	\centering
    \captionsetup{justification=centering}
	\caption{\textbf{image-AUROC comparison on MVTEC AD : MixedTeacher}}
	\footnotesize
    
	\setlength{\tabcolsep}{1.3pt}
	\renewcommand{\arraystretch}{1.3}
	\scalebox{0.8}{
	\begin{tabular}{|c|ccccc|c|}
			
		\hline
		{\centering \textbf{Category}} & 
		{\textbf{CutPaste}} &
		{\textbf{CFA}} &
		{\textbf{PatchCore}} &
		{\textbf{FastFlow}} &
		{\textbf{ReducedStudent}}&
		{\textbf{Ours}} \\
  
		\hline
		{carpet} &
		{\centering 100}  &
		{\centering 97.3} &
		{\centering 98.7}&
		{\centering 99.4}&
		{\centering \textbf{100}}&
		{\centering 99.8} \\

    	{tile} &
		{\centering 100}  &
		{\centering 99.4} &
		{\centering 98.7}&
		{\centering 100}&
		{\centering 98.7}&
		{\centering \textbf{100}} \\
		
		{grid} &
		{\centering 96.2}  &
		{\centering 99.2} &
		{\centering 98.2}&
		{\centering \textbf{100}}&
		{\centering 99.7} &
		{\centering 99.7}\\
		
		{wood} &
		{\centering 99.1}  &
		{\centering \textbf{99.7}} &
		{\centering 99.2}&
		{\centering 99.2}&
		{\centering 99.6}&
		{\centering 99.6} \\
		
		{leather} &
		{\centering 95.4}  &
		{\centering 100} &
		{\centering 100}&
		{\centering 99.9}&
		{\centering 99.7}&
		{\centering \textbf{100}}
		\\
		
		\hline
		{Mean} &
		{\centering 98.1}  &
		{\centering 99.1} &
		{\centering 99.0}&
		{\centering 99.7}&
		{\centering 99.5}&
		{\centering \textbf{99.8}} \\
		
	    \hline
 		
	\end{tabular} \label{table:Table4}
	}
\end{table}

Our method is the new state of the art texture anomaly detection on the MVTEC AD dataset.

\subsubsection{Anomaly localisation}

Even though anomaly localisation was not our main purpose, our approach uses EfficientNet-b0 with the objective of making the location more precise. To this end, we present in table \ref{table:Table5} and  table \ref{table:Table6}, our anomaly location results on textures from MVTEC AD dataset and BTAD respectively and we compare these results to the SOTA methods.

\begin{table}[h]
	\centering
    \captionsetup{justification=centering}
	\caption{\textbf{Pixel-AUROC comparison on MVTEC AD : MixedTeacher}}
	\footnotesize
	\setlength{\tabcolsep}{4pt}
	\renewcommand{\arraystretch}{1.3}
	\scalebox{1}{
	\begin{tabular}{|c|ccc|c|}
			
		\hline
		{\centering \textbf{Category}} & 
		{\textbf{CutPaste}} &
		{\textbf{PatchCore}} &
		{\textbf{FastFlow}} &
		{\textbf{Ours}} \\
	
		\hline
		{carpet} &
		{\centering 98.3}  &
		{\centering 98.9}&
		{\centering 99.1}&
		{\centering 99.0} \\

    	{tile} &
		{\centering 90.5}  &
		{\centering 95.6}&
		{\centering 96.6}&
		{\centering 95.9} \\
		
		{grid} &
		{\centering 97.5}  &
		{\centering 98.7}&
		{\centering 99.2}&
		{\centering 97.5}\\
		
		{wood} &
		{\centering 95.5}  &
		{\centering 95}&
		{\centering 94.1}&
		{\centering 94.9} \\
		
		{leather} &
		{\centering 99.5}  &
		{\centering 99.3}&
		{\centering 99.6}&
		{\centering 99.4}
		\\
		
		\hline
		{Mean} &
		{\centering 96.2} &
		{\centering 97.5}&
		{\centering 97.7}&
		{\centering 97.3} \\
		
	    \hline
 		
	\end{tabular} \label{table:Table5}
	}
\end{table}

\begin{table}[h]
	\centering
    \captionsetup{justification=centering}
	\caption{\textbf{Image-AUROC comparison on BTAD: MixedTeacher}}
	\footnotesize
	\setlength{\tabcolsep}{13pt}
	\renewcommand{\arraystretch}{1.3}
	\scalebox{1}{
	\begin{tabular}{|c|cc|}
			
		\hline
		{\centering \textbf{Category}} & 
		{\textbf{FastFlow}} &
		{\textbf{Ours}} \\
	
		\hline
		{1 (wood from btad)} &
		{\centering 96.0}  &
		{\centering \textbf{97.0}} \\
	    \hline
 		
	\end{tabular} \label{table:Table6}
	}
\end{table}

\subsubsection{Inference time results}
In terms of inference speed, our MixedTeacher is 3x slower than the reduced student since it used two teacher networks and a more complex student architecture.

\section{CONCLUSION}
In this paper, we proposed two methods for efficient unsupervised anomaly detection using the principle of knowledge distillation applied to unsupervised anomaly training. Both methods offer different benefits. The reduced student proposes a high speed texture anomaly detector with an AUROC performance close to the state of the art, this method is to be used in situations where inference time is the most important priority (mobile device, low computational power, cost efficiency). The MixedTeacher propose the highest actual performance on anomaly detection with a localisation close to the state of the art on the MVTEC AD textures with still a fast inference. This method is to be used in situations where performance is the priority and the computational power is big enough (monitoring server etc ...)

\small
\printbibliography
\end{document}